\def\year{2000}\relax
\documentclass[letterpaper]{article} 
\usepackage{aaai21}  
\usepackage{times}  
\usepackage{helvet} 
\usepackage{courier}  
\usepackage[hyphens]{url}  
\usepackage{graphicx} 
\usepackage{natbib}  
\usepackage{caption} 
\usepackage{multirow}
\usepackage{amsmath,amssymb}
\usepackage[switch]{lineno}  %
\title{Teacher-Critical Training Strategies for Image Captioning}
\author {
    Yiqing Huang\textsuperscript{\rm 1}, Jiansheng Chen\textsuperscript{\rm 1}\thanks{Corresponding author}\\ 
}
\affiliations {
    \textsuperscript{\rm 1}Department of Electronic Engineering, Tsinghua University
}
\begin{document}
\maketitle

\begin{abstract}
Existing image captioning models are usually trained by cross-entropy~(XE) loss and reinforcement learning~(RL), which set ground-truth words as \textbf{hard} targets and force the captioning model to learn from them.
However, the widely adopted training strategies suffer from misalignment in XE training and inappropriate reward assignment in RL training. 
To tackle these problems, we introduce a teacher model that serves as a bridge between the ground-truth caption and the caption model by generating some easier-to-learn word proposals as \textbf{soft} targets.
The teacher model is constructed by incorporating the ground-truth image attributes into the baseline caption model.
To effectively learn from the teacher model, we propose Teacher-Critical Training Strategies~(TCTS) for both XE and RL training to facilitate better learning processes for the caption model.
Experimental evaluations of several widely adopted caption models on the benchmark MSCOCO dataset show the proposed TCTS comprehensively enhances most evaluation metrics, especially the Bleu and Rouge-L scores, in both training stages.
TCTS is able to achieve to-date the best published single model Bleu-4 and Rouge-L performances of 40.2\% and 59.4\% on the MSCOCO Karpathy test split.
Our codes and pre-trained models will be open-sourced.
\end{abstract}

\section{Introduction}
\noindent 
Image captioning aims to automatically generate captions for images in natural language, which is of great significance in both Computer Vision and Natural Language Processing fields.
This challenging task facilitates lots of practical applications such as human-machine interaction and content-based image retrieval.
In general, image captioning models utilize CNN to encode the visual features of the image and leverage the LSTM~\cite{hochreiter1997long} or the Transformer~\cite{vaswani2017attention} as the language decoder to generate the captions.
Recently, the attention-based encoder-decoder framework~\cite{vinyals2015show} becomes prevalent in image captioning.
Generally speaking, different attention mechanisms focus on attending to different kinds of information.
Visual attention~\cite{xu2015show,anderson2018bottom} exploits the extracted spatial feature or objects feature to effectively utilize the visual information.
Semantic attention~\cite{you2016image,huang2020image} and adaptive attention focus more on linguistic information such as image attributes and language features.
Recently, X-linear attention~\cite{pan2020x} is proposed to model higher level information interactions.

\begin{figure}[tp]
	\centering
	\includegraphics[width=\linewidth]{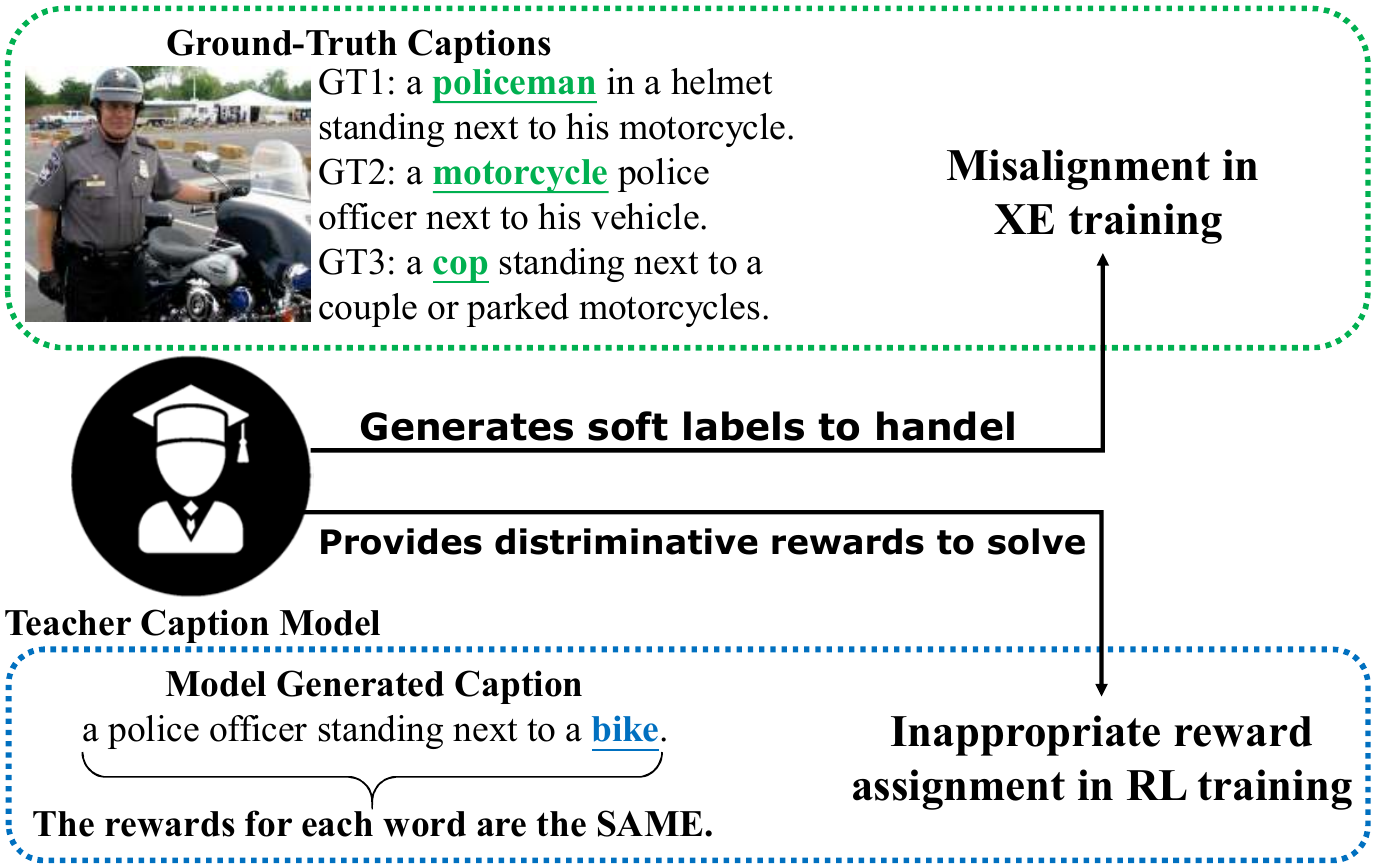}
	\caption{We propose to leverage a teacher caption model to solve the problems in current cross-entropy~(XE) and reinforcement learning~(RL) training processes.
	}
	\label{fig:fir}
\end{figure}
Although current image captioning methods achieve remarkable performances, the widely adopted model training strategies are far from satisfactory.
The captioning models are usually trained in two stages, namely the cross-entropy~(XE) training stage and reinforcement learning~(RL) training stage.
In the XE training stage, the model is trained to maximize the probability of the subsequent word given the previous ground-truth word.
However, we notice that about \textbf{98\%} percent of the images in the MSCOCO~\cite{chen2015microsoft} training set suffers from \textit{misalignment}.
As is shown in the green circle in Figure~\ref{fig:fir}, at the second time step, the model is fed with the word \textit{`a'} and is imposed to predict different words such as \textit{`policeman'}, \textit{`motorcycle'}, and \textit{`cop'} simultaneously.
Such supervision would confuse the model in maximizing the probability of which word.
In the RL stage, The widely adopted self-critical sequence training~(SCST)~\cite{Rennie2017SelfCriticalST} computes the CIDEr score to form the reward for the whole caption instead of for each word, which means that each word in the generated caption is assigned with the same reward. 
As shown in the blue circle in Figure~\ref{fig:fir}, the generated caption is semantically close to the ground-truth captions and it should be assigned with a relatively high reward.
Thus, the inaccurate word \textit{`bike'} also receives a high reward according to SCST.
Consequently, it is hard for the model to identify that the word \textit{`bike'} is not so appropriate and correct this mistake.
Therefore, we propose to leverage a teacher model, which achieves much better performance than current image captioning models, to guide the training process of the caption model~(or denoted as the student model).

Instead of using larger models with much more parameters and computation as the teacher~\cite{zhang2020object}, we propose to utilize attributes to train the teacher model.
The image attributes are the most salient words describing the image, thus incorporating the ground-truth attributes enables the teacher model to generate correct keywords in the caption.
Several previous works~\cite{yao2017boosting,li2019entangled} show that leveraging the ground-truth attributes boosts the performance of the captioning model by nearly 50\%.
The teacher model equipped with ground-truth attributes is applicable to guiding the student caption model in both XE and RL stages with our proposed Teacher-Critical Training Strategies~(TCTS).
In the XE stage, apart from utilizing the ground-truth captions as hard labels, we propose to leverage the teacher model to generate word probabilities as soft labels.
As the teacher model generates the same probability distribution for the same input, the misalignment in the XE training can be mitigated.
In the RL stage, we adopt the caption generated by the teacher, namely the teacher caption, as a sequence-level soft label.
Although the teacher caption is not necessarily more accurate than the ground-truth caption, its tone is very similar to the model generated caption.
Thus, we can effectively utilize the teacher caption to discriminate the appropriate words and inaccurate words inside the generated caption and adjust their rewards.
For instance, the reward for the word \textit{`bike'} in Figure~\ref{fig:fir} will be properly lowered.
Therefore, the student model will be able to recognize the inaccurate word and predicts a more appropriate word instead.

We evaluate both XE and RL performances of our proposed training strategies on the benchmark MSCOCO dataset.
To fairly compare and convincingly validate the effectiveness of our methods, we incorporate TCTS into several baseline models such as Att2in~\shortcite{Rennie2017SelfCriticalST}, BUTD~\shortcite{anderson2018bottom}, X-LAN and X-Trans~\shortcite{pan2020x}. 
Experimental results show that our approaches achieve comprehensive improvements, especially in terms of the Bleu scores, over these baseline models.
Our methods achieve to-date the best single model Bleu-4 scores of 40.2\% in the offline evaluation and 39.0\%/70.3\% in the online evaluation.
The main contributions of our works are as follows:
\begin{itemize}
	\item We propose Teacher-Critical Training Strategies~(TCTS) to improve the unreasonable supervision mechanisms in current training strategies to achieve state-of-the-art Bleu-4 and Rouge-L performances.
	\item TCTS provides additional soft labels to mitigate the misalignment in the XE training stage.
	\item In the RL stage, TCTS adjusts the rewards for appropriate words and inaccurate words to facilitate more reasonable reward assignment.
\end{itemize}
\begin{figure*}[htbp]
	\centering
	\includegraphics[width=\linewidth]{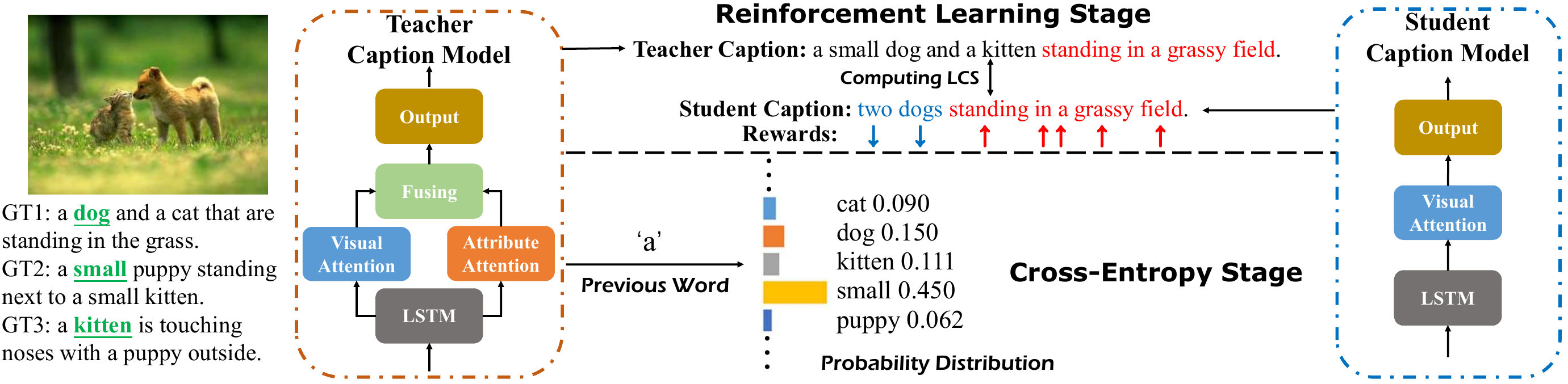}
	\caption{The overall framework of our proposal.
		We utilize the ground-truth image attribute to construct the teacher caption model.
		To effectively utilize attribute information, we implement the attribute attention module and fuse the attended attribute information and visual information.
		The teacher model is applicable to guide the training process of the student model in both XE stage and the RL stage.
		In the XE stage, the teacher model generates the same probability distribution with identical previous input word \textit{`a'} as soft labels to alleviate the misalignment in the ground-truth captions~(shown in green).
		In the RL stage, we compute the Longest Common Subsequence~(LCS, shown in red) between the teacher caption and the student caption to adjust the rewards for appropriate words and inaccurate words~(shown by red and blue arrows).
	}
	\label{fig:overall}
\end{figure*}
\section{Related Works}
\textbf{Image Captioning}~
Most modern image captioning methods~\cite{xu2015show,you2016image,lu2017knowing,anderson2018bottom, huang2019attention, pan2020x} encode image through CNNs and then decode it with RNNs or the Transformer~\cite{vaswani2017attention}.
Multiple attention mechanisms are proposed to enhance image captioning.
In particular, Xu~\textit{et al.}~\shortcite{xu2015show} introduced soft and hard attention mechanisms to select the most relevant image regions in word generation. 
You~\textit{et al.}~\shortcite{you2016image} proposed semantic attention to attend to the most salient words, namely image attributes, in the image in the language decoder.
Anderson~\textit{et al.}~\shortcite{anderson2018bottom} leveraged the Faster R-CNN~\cite{renNIPS15fasterrcnn} to extract more explicit object features and predicted the image captions via bottom-up and top-down attention. 
Recently, Pan~\textit{et al.}~\shortcite{pan2020x} designed an X-Linear attention block to capture higher-order interactions between the visual features to achieve state-of-the-art performance.

\noindent\textbf{Training Strategies}~The caption models are always trained in two stages.
In the first stage, the model is trained to predict the next token given previous ground-truth words under the cross-entropy~(XE) loss.
Zhang~\textit{et al.}~\shortcite{zhang2020object} introduced a teacher-recommended method to distill knowledge from an external language model.
Although they also proposed to leverage soft label in XE training with a teacher model, taking a pure language model as the teacher may suffer from modality bias since the visual feature is significant in visual captioning.
Thus, we propose to directly take an image captioning model as the teacher model to more effectively mitigate the misalignment in XE training.
Moreover, while their method requires additional data to train the teacher model, our teacher model is constructed by additionally taking the ground-truth image attribute, which can be easily obtained from the ground-truth captions.

In the second stage, reinforcement learning~(RL) is introduced to directly optimize the evaluation metrics such as CIDEr and Bleu.
Rennie~\textit{et al.}~\shortcite{Rennie2017SelfCriticalST} introduced the self-critical sequence training~(SCST) method which obtained a reward by taking the current model under the inference algorithm as the baseline.
Discriminative reward for each word is not considered under this training strategy.
To remedy this, Zhang~\textit{et al.}~\shortcite{zhang2017actor} used another RNN to predict the state value function for different words.
However, the value they computed is not directly related to the evaluation scores, which introduces estimation bias.
Recently, Gao~\textit{et al.}~\shortcite{gao2019self} proposed N-step SCST to estimate the per-token value by additionally generating several captions in the training process.
Although their method utilizes the differences of CIDEr scores to assess the words' value, they omit to evaluate the quality of the whole caption.
Thus, the model may fall into local maxima where each word is plausible but the whole caption is less satisfying.
Moreover, they suffer from relatively high computation cost, such as inference and computing CIDEr score several times in one iteration.
While these previous methods criticize the captioning model by itself or a value network, we leverage a model that achieves much higher performance than current captioning models, as the teacher to criticize the caption model from a more experienced perspective.
Moreover, the major additional computational cost of our method is only to search the longest common subsequence at each training step which can be performed using a polynomial time complexity algorithm.

\section{Method}
Figure~\ref{fig:overall} demonstrates the overview of our proposed method.
We firstly construct a teacher model that exploits both visual features and the ground-truth image attributes.
The teacher model generates easy-to-learn word proposals in both XE and RL training stages to alleviate the problems in current training strategies.
The implementation details of the teacher model and the Teacher-Critical Training Strategies~(TCTS) are elaborated in this section.
\subsection{Constructing the Teacher Model}
In this work, we adopt the X-LAN~\cite{pan2020x} model as the backbone.
This model consists of an X-linear attention based visual feature encoder and an X-linear attention module based LSTM decoder.
While the vanilla X-LAN model only exploits the visual feature of each image, we additionally leverage another encoder to encode the embedding of ground-truth image attributes as is shown in the brown box in Figure~\ref{fig:overall}.
Denoting the attended visual feature and image attributes as $\boldsymbol{\hat{v}}_t$ and $\boldsymbol{\hat{a}}_t$ respectively, we fuse them with the output of the LSTM layer $\boldsymbol{h}_t$ as in Eq~\ref{eq:fuse_v}-\ref{eq:fuse}.
Here $\boldsymbol{\alpha}_t$ and $\boldsymbol{\beta}_t$, each element in which is in range $[0,1]$, are the fusion weights; the $\mathbf{W}$s are trainable parameters.
\begin{equation}
	\boldsymbol{\alpha}_t=\mathit{Sigmoid}(\mathit{Concat}(\boldsymbol{h}_t;\enspace\boldsymbol{\hat{v}}_t)\mathbf{W}^{v}) \label{eq:fuse_v}
\end{equation}
\begin{equation}
	\boldsymbol{\beta}_t=\mathit{Sigmoid}(\mathit{Concat}(\boldsymbol{h}_t;\enspace\boldsymbol{\hat{a}}_t)\mathbf{W}^{a}) \label{eq:fuse_a}
\end{equation}
\begin{equation}
	\boldsymbol{\hat{f}}_t= \boldsymbol{\alpha}_t*\boldsymbol{\hat{v}}_t + \boldsymbol{\beta}_t*\boldsymbol{\hat{a}}_t\label{eq:fuse}
\end{equation}
The fused feature $\boldsymbol{\hat{f}}_t$ is then concatenated with $\boldsymbol{h}_t$ and sent to the Gated Linear Unit~(GLU) layer to generate the context as in Eq~\ref{eq:glu}.
The context is finally input to the linear layer to generate the probability distribution of each word in the vocabulary shown in Eq~\ref{eq:ptword}.
\begin{equation}
\boldsymbol{c}_t=\mathit{GLU}([\boldsymbol{h}_t;\enspace\boldsymbol{\hat{f}}_t]\mathbf{W}^{glu}) \label{eq:glu}
\end{equation}
\begin{equation}
\boldsymbol{q}_t=\mathit{Softmax}(\boldsymbol{c}_t\mathbf{W}^{p})\label{eq:ptword}
\end{equation}

Taking the ground-truth attribute as privilege knowledge, the teacher model is capable of generating correct key words when narrating the captions and forms applicable guidance for the student in both XE and RL training stages.

\subsection{TCTS for XE Training}
Denoting the ground-truth caption as $\{y_1,\enspace y_2,...,\enspace y_T\}$, the caption model is forced to maximize the probability of these ground-truth words in the XE training stage.
Specifically, the loss function is computed as in Eq~\ref{eq:xe}, where $\boldsymbol{\theta}$ is the trainable parameter in the caption model and $\boldsymbol{p}_t$ is the word probability distribution predicted by the caption model.
\begin{equation}
  \mathit{L}^{XE}(\boldsymbol{\theta})=-\frac{1}{T}\Sigma_{t=1}^{T}\delta(y_t)*\mathit{log}\boldsymbol{p}_t \label{eq:xe}
\end{equation}
It can be noticed that only the probability of the ground-truth word is directly optimized since $\delta(y_t)$ is a one-hot vector which equals to 1 only at the $y_t^{th}$ position.
This constraint introduces misalignment since some corresponding words, as shown in the green words in Figure~\ref{fig:overall}, such as \textit{`dog'}, \textit{`small'}, and \textit{`kitten'} are simultaneously forced to reach the top probability given the same previous word \textit{`a'}.
Alternatively, we adopt the word probability generated by the teacher model $\boldsymbol{q}_t$ in the forward pass as the soft label as is shown in the bottom of Figure~\ref{fig:overall}.
We additionally compute the Kullback-Leibler~(KL) divergence between the distribution of the teacher model and the student model to soften the loss function.
The loss of KL divergence is formulated in Eq~\ref{eq:kl}, where $K$ is the vocabulary size.
As the teacher model generates the same probability distribution with identical inputs, the misalignment in the XE training can be mitigated.
Thus, the student model is taught to generate a more reasonable probability distribution rather than maximize the probability of ground-truth word at each time step.
\begin{equation}
	\mathit{L}^{KL}(\boldsymbol{\theta})=-\frac{1}{T}\Sigma_{t=1}^{T}\Sigma_{i=1}^K\boldsymbol{q}_t^{i}*\mathit{log}\frac{\boldsymbol{p}^i_t}{\boldsymbol{q}_t^{i}} \label{eq:kl}
\end{equation}
The KL divergence loss is finally summed with XE loss by the weight of $\lambda_1$ to form the teacher-critical XE loss in Eq~\ref{eq:xe+tea}.
\begin{equation}
	\mathit{L}^{XE}_{TCTS}(\boldsymbol{\theta})=\mathit{L}^{XE}(\boldsymbol{\theta})+\lambda_1*\mathit{L}^{KL}(\boldsymbol{\theta})\label{eq:xe+tea}
\end{equation}

\subsection{TCTS for RL Training}
Only optimizing the captioning model with the XE loss would lead to exposure bias.
The model is fed with the ground-truth word in the training process but is fed with the previous generated word in the inference mode.
More importantly, the evaluation metric are not directly optimized in the training process. 
Consequently, several reinforcement learning~(RL) strategies are proposed to remedy these inconsistency.
We first briefly introduce the general idea of implementing RL in image captioning.
The caption model, parameterized by $\boldsymbol{\theta}$, can be viewed as an agent who receives a reward $\mathbb{R}$ after it takes an action $\mathbb{A}$~(generates a caption $w^s$) against the current environments (input features).
The goal of reinforcement learning is to minimize the negative expect reward as in Eq~\ref{eq:reward}, where $p_\theta$ is the policy defined by the parameter $\boldsymbol{\theta}$.
In practice, the reward is estimated with a single sample from $p_\theta$. 
Therefore, the gradient of reinforcement learning is formulated in Eq~\ref{eq:rl_loss}.
\begin{equation}
	\mathit{L}^{RL}(\boldsymbol{\theta})=-\mathbb{E}_{w^s\sim p_\theta}[\mathbb{R}(w^s)] \label{eq:reward}
\end{equation}
\begin{equation}
\nabla_{\boldsymbol{\theta}}\mathit{L}^{RL}(\boldsymbol{\theta})=-\mathbb{R}(w^s)\nabla_{\boldsymbol{\theta}}\mathit{log}p_\theta(w^s),\enspace w^s\sim p_\theta \label{eq:rl_loss}
\end{equation}
In the widely-adopted Self-Critical Sequence Training~(SCST)~\cite{Rennie2017SelfCriticalST}, the reward is the CIDEr score of the sampled caption minus the CIDEr score of the greedy decoded caption as in Eq~\ref{eq:scst}.
\begin{equation}
\mathbb{R}(w^s)^{SCST}=\mathit{CIDEr}(w^s)-\mathit{CIDEr}(w^{gre})\label{eq:scst}
\end{equation}
Note that this reward is computed for the whole caption, which means that the reward of all the words in the sampled caption are the same.
However, as shown in the top of Figure~\ref{fig:overall}, the student model correctly narrates \textit{`standing'} but erroneously states \textit{`two dogs'}.
It is not reasonable to assign \textit{`standing'} and \textit{`two dogs'} the same reward.

Consequently, we additionally leverage a teacher model to criticize the caption sampled from the student model to assign more reasonable weights to each word.
Practically, the teacher greedy decodes a caption denoted as $w^{tea}=[w^{tea}_1,\enspace w^{tea}_2,..,w^{tea}_T]$.
As the teacher model is trained with the ground-truth attributes, most keywords in the teacher captions are correct.
Thus, the teacher caption is well formulated and can be leveraged to guide the student model
We then search the Longest Common Subsequence~(LCS) between the teacher caption and the student caption shown at the top of Figure~\ref{fig:overall}.
A subsequence is a sequence that appears in the same relative order, but not necessarily contiguous.
The LCS problem is a classic computer science problem that aims at finding the longest subsequence present in both given sequences.
The words inside the LCS are assumed to be the \textbf{appropriate} words that should be assigned with more rewards.
The rest words are more likely to be \textbf{inaccurate}, therefore their rewards should be decreased.
Suppose that there are $n$ appropriate words and $m$ inaccurate words in the student caption, we compute the normalized CIDEr score of the teacher caption to modify the original reward in Eq~\ref{eq:scst}.
The additional rewards for the appropriate words and the inaccurate words are formulated in Eq~\ref{eq:pos}-\ref{eq:neg}.
\begin{equation}
	\eta=\frac{n-m}{n+m}\mathit{CIDEr}(w^{tea}) \label{eq:pos}
\end{equation}
\begin{equation}
	\mathbb{R}(appr) = \mathit{CIDEr}(w^{tea})-\eta 
\end{equation}
\begin{equation}
	\mathbb{R}(inac) = -\mathit{CIDEr}(w^{tea})-\eta \label{eq:neg}
\end{equation}
The additional rewards are added to the original reward with the weight of $\lambda_2$ in Eq~\ref{eq:final_reward} to form the teacher-critical reward for each appropriate and inaccurate word respectively as shown by the red and blue arrows in Figure~\ref{fig:overall}.
\begin{equation}
	\mathbb{R}(w^s)^{TCTS} = \mathbb{R}(w^s)^{SCST}+\lambda_2\mathbb{R}(appr/inac) \label{eq:final_reward}
\end{equation}
Note that the sum of the normalized additional reward of a caption equals zero, therefore, taking the teacher-critical rewards does not violate the goal of optimizing CIDEr score in the reinforcement learning stage but encourages the student model to generate more appropriate words and replace the inaccurate words.

It should be noticed that teacher captions may not be necessarily better than the ground-truth captions semantically.
However, since they are generated by a \textbf{model} instead of \textbf{human beings}, they better follow the tone of how a deep model may describe images.
Therefore, they are probably more achievable supervisions for the student model compared to the human labelled ground-truth.
Actually, a teacher caption can be viewed as sequence-level soft label that effectively bridges the gap between the student model and the human-labeled ground-truth captions.
Detailed comparisons will be elaborated in the quantitative analysis.

\begin{table}[tbp]
	\small
	\centering
	\resizebox{0.46\textwidth}{13mm}{
		\begin{tabular}{@{\extracolsep{\fill}}c|cccccccc}
			\hline
			Methods&\multicolumn{8}{c}{Cross-Entropy Loss}\\
			\hline
			Metric&B-1&B-2&B-3&B-4&M&R&C&S\\
			\hline\hline
			BUTD~\shortcite{anderson2018bottom}&77.2&-&-&36.2&27.0&56.4&113.5&20.3\\
			+TCTS (ours)&77.5&61.7&47.6&37.0&28.1&57.3&116.6&21.2\\ \hline
			X-Trans~\shortcite{pan2020x}&77.3&61.5&47.8&37.0&28.7&57.5&120.0&21.8\\
			+TCTS (ours)&77.6&61.9&48.3&37.5&28.7&57.7&120.8&21.8\\ \hline
			X-LAN~\shortcite{pan2020x}&78.0&62.3&48.9&38.2&28.8&58.0&122.0&21.9\\
			+TCTS (ours)&\textbf{78.3}&\textbf{62.9}&\textbf{49.3}&\textbf{38.3}&\textbf{28.9}&\textbf{58.2}&\textbf{122.3}&\textbf{22.0}\\			\hline
	\end{tabular}}
	\caption{Single model offline performances~(\%) of various methods on MSCOCO trained by cross-entropy loss only, where B-N, M, R, C and S are short for Bleu-N, Meteor, Rouge-L, CIDEr and SPICE scores. The best score in each column is marked in \textbf{boldface}.}
	\label{tab:offline-xe}
\end{table}
\begin{table*}[htbp]
	\small	
	\centering
	\begin{tabular}{@{\extracolsep{\fill}}c|cccccccc}
		\hline
		Methods&\multicolumn{8}{c}{Reinforcement Learning}\\
		\hline
		Metric&Bleu-1&Bleu-2&Bleu-3&Bleu-4&Meteor&Rouge-L&CIDEr&SPICE\\
		\hline\hline
		GCN-LSTM~\cite{yao2018exploring}&80.5&-&-&38.2&28.5&58.3&127.6&22.0\\
		LBPF~\cite{qin2019look}&80.5&-&-&38.3&28.5&58.4&127.6&22.0\\
		SGAE~\cite{yang2019auto}&80.8&-&-&38.4&28.4&58.6&127.8&22.1\\
		AoA~\cite{huang2019attention}&80.2&-&-&38.9&29.2&58.8&129.8&22.4\\
		MAD+SAP~\cite{huang2020image}&-&-&-&38.6&28.7&58.5&128.8&22.2\\
		M2Trans~\cite{cornia2020meshed}&80.8&-&-&39.1&29.2&58.6&131.2&22.6\\
		\hline
		BUTD~\cite{anderson2018bottom}&79.8&-&-&36.3&27.7&56.9&120.1&21.4\\
		+TCTS~(ours)&80.3&65.0&50.7&38.8&28.6&58.7&126.4&22.1\\
		\hline						X-Trans~\cite{pan2020x}&80.9&65.8&51.5&39.7&\textbf{29.5}&59.1&\textbf{132.8}&23.4\\
		X-Trans*&81.0&65.8&51.5&39.6&29.4&59.0&132.0&23.4\\
		+TCTS~(ours)&\textbf{81.2}&\textbf{66.1}&51.9&40.1&\textbf{29.5}&59.3&132.3&\textbf{23.5}\\ \hline
		X-LAN~\cite{pan2020x}&80.8&65.6&51.4&39.5&\textbf{29.5}&59.2&132.0&23.4\\
		X-LAN*&80.7&65.5&51.2&39.3&\textbf{29.5}&59.0&131.7&23.3\\
		+TCTS~(ours)&81.0&\textbf{66.1}&\textbf{52.0}&\textbf{40.2}&\textbf{29.5}&\textbf{59.4}&132.2&23.4\\ \hline
	\end{tabular}
	\caption{Single model offline performances~(\%) of various methods on MSCOCO trained by reinforcement learning.}
	\label{tab:offline-rl}
\end{table*}
\section{Expriments}
\subsection{Experimental Settings}
We evaluate our model on the MSCOCO captioning dataset~\cite{chen2015microsoft}.
Words that appear in the training set for over $5$ times are selected to form a vocabulary of the size $K$=$9487$.
We follow the widely adopted Karpathy's data split~\cite{karpathy2015deep} in offline evaluation.
We utilize the ResNet-101 object features released in BUTD~\cite{anderson2018bottom} for image captioning.
For fair comparisons with the SOTA X-LAN~\cite{pan2020x} and re-productivity, the explicit model sizes~(\textbf{including the random seeds}) and learning rate schedules are set identically to their open-source codes.
Our models are trained $50$ epochs under XE+TCTS loss and another $50$ epochs under SCST+TCTS with a mini-batch size of $40$ and $20$ respectively.
To evaluate the image captioning performance, the following metrics are used: Bleu~\shortcite{papineni2002bleu}, Meteor~\shortcite{Denkowski2014MeteorUL}, Rouge-L~\shortcite{lin2004rouge},
CIDEr~\shortcite{vedantam2015cider}, and SPICE~\shortcite{anderson2016spice}.
Similar to~\cite{fang2015captions,gan2017semantic}, the most frequent 1000 words are selected to form the attribute vocabulary in the teacher model.
The teacher model is fixed after XE and RL training.
The weights of TCTS in XE and RL training are set to $\lambda_1=0.2$ and $\lambda_2=0.02$ respectively.
\subsection{Performance Evaluation}
\noindent\textbf{Offline Evaluation} Table~\ref{tab:offline-xe} reports the performances of our proposed TCTS in terms of XE training on several baseline models.
It can be noticed that with the misalignment mitigated, comprehensive improvement is reached for these baseline models.
In Table~\ref{tab:offline-rl} we present the gain of incorporating TCTS in the RL stage.
Note that the X-LAN and X-Trans~\cite{pan2020x} models are trained with a batch size of $40$.
However, we can only set the batch size to $20$ due to GPU constraint.
Consequently, we re-train these two models with a batch size of $20$ for fair comparisons.
The performances of the re-trained models are reported with \textit{`*'}.
In general, our proposed TCTS also consistently exhibits better performances than the baseline models in the RL stage, especially in terms of the Bleu and the Rouge-L scores.
This is because the proposed TCTS encourages student model to generate the captions that have longer LCS with the teacher captions.
As the Bleu-4 and Rouge-L score is closely related to the LCS and identical 4-grams between the model generated caption and the ground-truth, higher scores implies that the semantic quality of teacher caption is high enough to guide the student model in the training process.

Comparing with other recent image captioning methods in Table~\ref{tab:offline-rl}, we notice that the SGAE~\cite{yang2019auto} and the BUTD+TCTS model show comparable performances.
While SGAE adopts the BUTD model as the backbone and additionally incorporates scene graph features, the BUTD+TCTS model outperforms SGAE in terms of Bleu-4, Meteor, and Rouge-L scores without taking in any additional inputs.
Utilizing the LSTM based X-LAN teacher to guide X-Trans* leads to relatively less improvement, perhaps this is due to the fact that the LSTM is less powerful than the Transformer in the RL stage.
Putting Table~\ref{tab:offline-xe} and Table~\ref{tab:offline-rl} together, we can witness that directly incorporating the widely adopted SCST~\cite{Rennie2017SelfCriticalST} improves 1.1\% Bleu-4 and 1.0\% Rouge-L scores on the X-LAN* model.
However, the Bleu-4 and Rouge-L scores enjoy 1.9\% and 1.2\% improvements respectively with the the guidance from the teacher model, which suggests that our proposed RL method is more effective than SCST on the SOTA captioning model.
In general, utilizing our propsed TCTS reaches the best Bleu and Rouge-L scores among all compared methods. 
The performance comparisons again verify the effectiveness of adopting TCTS in the RL stage.

\noindent\textbf{Online Evaluation} Table~\ref{tab:online} reports the performance gain of utilizing our proposed TCTS on the X-LAN* model in the RL stage on the official MSCOCO evaluation server.
Compared with other methods that utilize an ensemble of multiple captioning models, we can see that our single model achieves very competitive performance.
Similar to the offline evaluation, we reach the best Bleu-2, Bleu-3, Bleu-4 and Rouge-L scores among all the compared methods.

\begin{table*}[tbp]
	\centering
		\begin{tabular}{@{\extracolsep{\fill}}ccccccccccccccc}   
			\hline
			\multirow{2}*{Methods}&\multicolumn{2}{c}{Bleu-1}&\multicolumn{2}{c}{Bleu-2}&\multicolumn{2}{c}{Bleu-3}&\multicolumn{2}{c}{Bleu-4}&\multicolumn{2}{c}{Meteor}&\multicolumn{2}{c}{Rouge-L}&\multicolumn{2}{c}{CIDEr}\\
			\cline{2-15}
			\quad&c5&c40&c5&c40&c5&c40&c5&c40&c5&c40&c5&c40&c5&c40\\
			\hline\hline
			SCST$^\Sigma$~\shortcite{Rennie2017SelfCriticalST}&78.1&93.7&61.9&86.0&47.0&75.9&35.2&64.5&27.0&35.5&56.3&70.7&114.7&116.7\\
			LSTM-A$^\Sigma$~\shortcite{yao2017boosting}&78.7&93.7&62.7&86.7&47.6&76.5&35.6&65.2&27.0&35.4&56.4&70.5&116.0&118.0\\
			Up-Down$^\Sigma$~\shortcite{anderson2018bottom}&80.2&\textbf{95.2}&64.1&88.8&49.1&79.4&36.9&68.5&27.6&36.7&57.1&72.4&117.9&120.5\\
			CAVP~\shortcite{Zha2019ContextAwareVP}&80.1&94.9&64.7&88.8&50.0&79.7&37.9&69.0&28.1&37.0&58.2&73.1&121.6&123.8\\
			SGAE$^\Sigma$~\shortcite{yang2019auto}&-&-&-&-&-&-&38.5&69.7&28.2&37.2&58.6&73.6&123.8&126.5\\              
			MAD+SAP$^\Sigma$~\shortcite{huang2020image}&\textbf{80.5}&94.9&65.1&\textbf{89.1}&50.4&80.0&38.4&69.4&28.6&37.7&58.7&73.3&125.1&127.0\\ \hline
			X-LAN~\shortcite{pan2020x}&80.3&94.7&65.0&88.9&50.5&80.2&38.6&69.8&\textbf{29.1}&38.3&58.7&73.8&\textbf{125.4}&\textbf{127.3}\\
			X-LAN*&80.3&94.8&65.0&88.9&50.5&80.2&38.5&69.8&28.9&38.2&58.6&73.8&124.9&126.9\\
			+TCTS~(ours)&\textbf{80.5}&94.8&\textbf{65.3}&\textbf{89.1}&\textbf{50.9}&\textbf{80.5}&\textbf{39.0}&\textbf{70.3}&29.0&\textbf{38.4}&\textbf{58.9}&\textbf{74.0}&125.3&127.2\\ \hline
		\end{tabular}
	\caption{Performance (\%) on the online MSCOCO evaluation server. The methods using model ensemble are labeled with $^\Sigma$. }
	\label{tab:online}
\end{table*}

\begin{figure*}[ht]
	\centering
	\includegraphics[width=\linewidth]{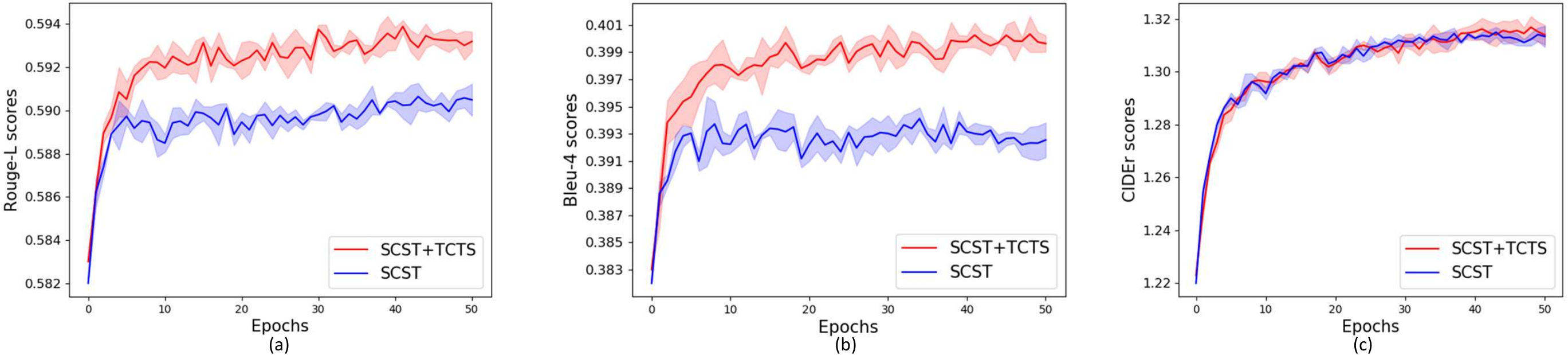}
	\caption{Performance comparisons of adopting TCTS over SCST in terms of Rouge-L, Bleu-4, and CIDEr scores. The horizotal axes are every epoch and the vertical axes are corresponding metrics on MSCOCO Karpathy test split}
	\label{fig:process}
\end{figure*}

\subsection{Quantitative Analysis}\label{sec:quantitative}
\noindent\textbf{The Quality of Teacher Caption}~The key to our proposed TCTS in the RL stage is that captions generated by the teacher model are better guidance for the caption model.
To quantitatively demonstrate the quality of the teacher caption, we utilize the teacher captions as the ground-truth captions to train the captioning model with the self-critical sequence training~\cite{Rennie2017SelfCriticalST}.
As the teacher model only generates one caption per image, we randomly select one human-labeled ground-truth caption per image to train another caption model for a fair comparison.
The comparison shown in Table~\ref{tab:compare-rl} suggests that using the teacher caption reaches higher a CIDEr score in the CIDEr optimization SCST than one ground-truth caption.
This is because the teacher caption is narrated by a model but not a human being.
Although the teacher caption is not necessarily better than the ground-truth caption in depicting the image, the language style of the teacher caption is closer to the model's way of narrating the captions.
Consequently, utilizing such captions as additional guidance in our proposed TCTS is beneficial for training caption models.
\begin{table}[tbp]
	\small	
	\centering
	\resizebox{0.47\textwidth}{7mm}{
		\begin{tabular}{@{\extracolsep{\fill}}c|cccccccc}
			\hline
			Methods&\multicolumn{8}{c}{Reinforcement Learning-SCST}\\
			\hline
			Metric&B-1&B-2&B-3&B-4&M&R&C&S\\
			\hline\hline
			OneGTCaption&\textbf{77.3}&\textbf{61.8}&48.2&\textbf{37.5}&29.6&58.4&125.4&22.4\\
			TeacherCap&77.2&\textbf{61.8}&\textbf{48.3}&37.3&\textbf{29.8}&\textbf{58.5}&\textbf{126.1}&\textbf{23.0}\\
			\hline						
	\end{tabular}}
	\caption{Single model offline performances~(\%) of adopting different ground-truth in SCST.}
	\label{tab:compare-rl}
\end{table}

\begin{figure*}[ht]
	\centering
	\includegraphics[width=\linewidth]{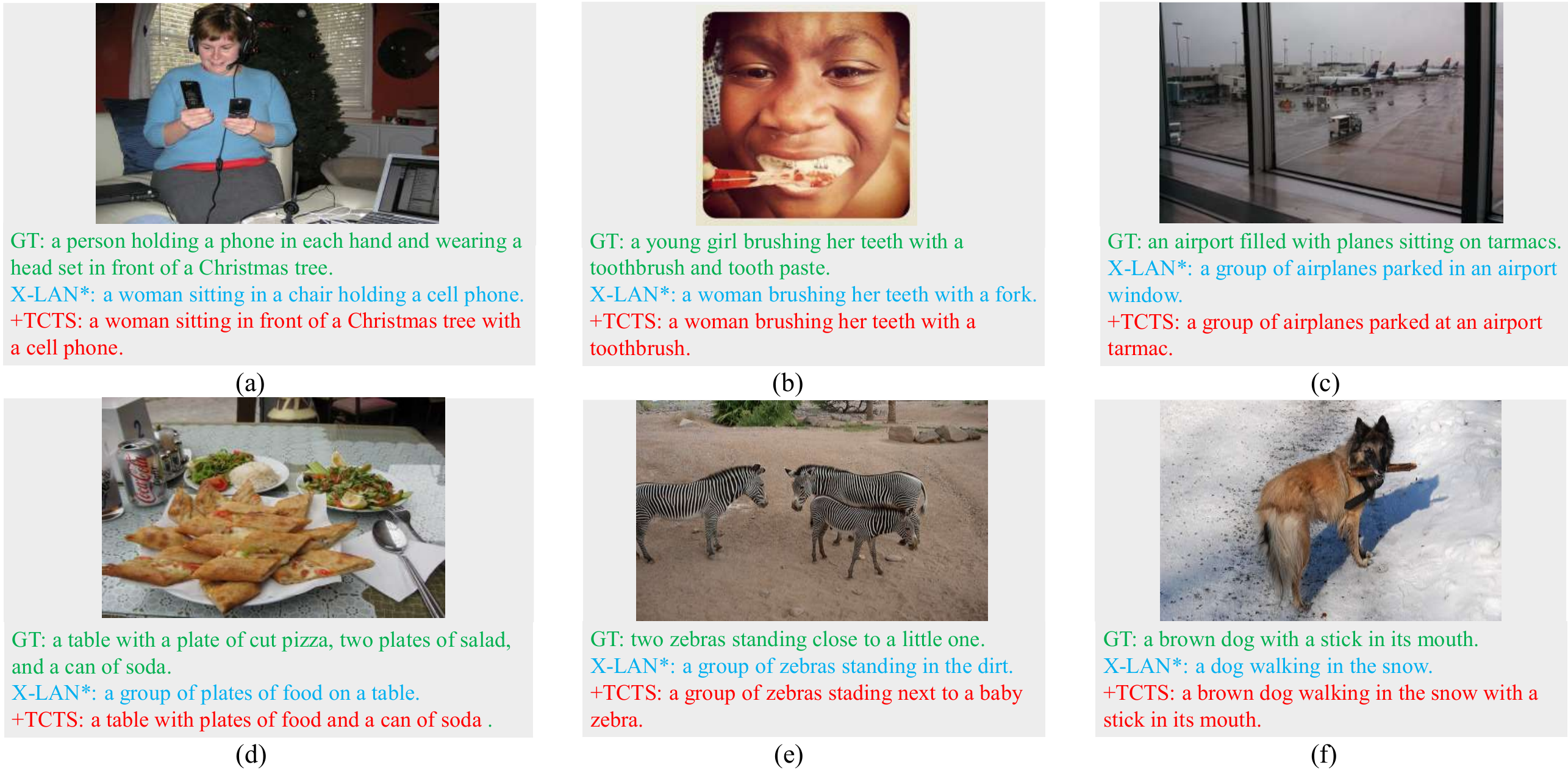}
	\caption{Qualitative results of our proposed TCTS (in red) compared with the ground-truth caption (in green) and the X-LAN* model trained under SCST (in blue).}
	\label{fig:qualitative}
\end{figure*}

\noindent\textbf{Evaluating Different Training Strategies}
We compare our method over several commonly used training strategies in Table \ref{tab:compare-st}.
For fair comparisons with the previous works, we re-implement the Att2in model~\cite{Rennie2017SelfCriticalST} which adopts $14\times14\times2048$ ResNet~\cite{he2016deep} feature in caption generation.
It can be witnessed that our proposed TCTS achieves comprehensively improvement in the XE training stage.
In the RL stage, we outperform the baseline SCST by a large margin.
Comparing with the more recent N-SCST~\cite{gao2019self} method, our model also achieves the best scores in all the metrics.
Specifically, our method enhances the CIDEr score for 3.4\% over the N-SCST method, which suggests that our method performs much better in the CIDEr optimization RL training process.

It is worth noticing that adopting TCTS yields more gain on the Att2in model than on the X-LAN* model.
Considering that Att2in model contains much fewer parameters than the X-LAN* model, we find that our methods is more effective on the lighter models.
Consequently, it is promising that our methods would contribute to the practical implementations of image captioning models.

\begin{table}[tbp]
	\small	
	\centering
	\resizebox{0.47\textwidth}{13mm}{
		\begin{tabular}{@{\extracolsep{\fill}}c|cccccccc}
			\hline
			Methods&B-1&B-2&B-3&B-4&M&R&C&S\\
			\hline\hline
			\multicolumn{9}{c}{Cross-Entropy Loss}\\ \hline
			XE~\shortcite{Rennie2017SelfCriticalST}&-&-&-&31.3&26.0&54.3&101.4&-\\
			+TCTS(ours)&75.3&59.1&45.2&34.4&26.5&55.4&107.5&19.7\\ \hline			
			\multicolumn{9}{c}{Reinforcement Learning}\\ \hline
			SCST~\shortcite{Rennie2017SelfCriticalST}&-&-&-&33.3&26.3&55.3&111.4&-\\
			N-SCST~\shortcite{gao2019self}&77.9&61.5&46.8&35.0&\textbf{26.9}&56.3&115.2&\textbf{20.4}\\
			+TCTS (ours)&\textbf{78.1}&\textbf{62.1}&\textbf{47.4}&\textbf{35.7}&\textbf{26.9}&\textbf{56.8}&\textbf{118.6}&\textbf{20.4}\\
			\hline						
	\end{tabular}}
	\caption{Single model offline performances~(\%) of adopting various training strategies on the Att2in model.}
	\label{tab:compare-st}
\end{table}

\noindent\textbf{Improvements Visualization}~To clearly demonstrate the gain of adopting our proposed teacher-critical training strategies over the widely adopted SCST, we compare the performance of three X-LAN* model and X-LAN*+TCTS model trained with the same three random seeds.
These models' average and standard derivation of Rouge-L, Bleu-4, and CIDEr scores on the MSCOCO test set are shown in Figure~\ref{fig:process}.
We find that our method shows remarkable improvement over the SCST baseline in terms of Rouge-L and Bleu-4 scores due to the discriminative reward computation strategy adopted in our TCTS.
Our training strategy encourages the student model to generate longer LCS with the teacher model and punish the words that are not in the LCS.
Thus the Rouge-L score and Bleu-4 score can be effectively enhanced.
Figure~\ref{fig:process}(c) shows that adopting TCTS even improves the CIDEr performance in latter epochs.
Consequently, additionally adopting our method not only does not violate the goal of optimizing the CIDEr score but also increases the CIDEr with the help of the teacher model.

\subsection{Qualitative Results}
Figure~\ref{fig:qualitative} shows some qualitative results of our proposed TCTS method
against the ground-truth caption and the X-LAN* model trained with SCST~\cite{Rennie2017SelfCriticalST}.
These images are chosen from the MSCOCO Karpathy test split.
We show three captions for each image, in which the green, blue, and red captions are the ground-truth and generated by X-LAN* and X-LAN*+TCTS respectively. 
Generally, it can be noticed that with the incorporation of TCTS, the student model can recognize more correct keywords for the image.
This is because the teacher model is trained with the ground-truth attributes, therefore it is capable of generating captions with correct keywords.
In the training process, the student model is encouraged to generate more appropriate keywords to prolong the LCS.
Such ability to recognize the objects and generate correct keywords can be generalized to the test set to assist the student model to generate precise and detailed captions.
As shown in Figure~\ref{fig:qualitative}(a)-(c), while the baseline X-LAN* model fails to depict the correct instances in the image, the X-LAN*+TCTS model is able to precisely recognize the \textit{`Christmas tree'}, the \textit{`toothbrush'}, and the \textit{`tarmacs'}.
We show that TCTS helps the student model to generate a more detailed caption in Figure~\ref{fig:qualitative}(d)-(f).
Our model narrates very similar to the ground-truth in Figure~\ref{fig:qualitative}(d), which means the teacher even helps the student to better learn the tone of the ground-truth to some extent.
In Figure~\ref{fig:qualitative}(f), our model even depicts the image more detailed than the ground-truth caption, which further verifies the effectiveness of the proposed TCTS.
\section{Conclusions}
In this paper, we propose Teacher-Critical Training Strategies~(TCTS) to effectively leverage the teacher model to improve the unreasonable supervision mechanisms in commonly adopted XE and RL training processes.
The experimental results on MSCOCO suggest that our method achieves comprehensive improvement in the XE stage and obtains remarkable enhancement in terms of Bleu and Rouge-L scores in the RL stage.
That our method achieves noticeable improvement over SCST on the light Att2in model suggests that our method is more applicable in the practical implementation of image captioning models.
Currently, our method is not capable of generate a per-token reward for each word in the student caption.
We leave it as the future work that developing a more advanced teacher-critical per-token reward assignment strategy to further boost the CIDEr score.

{\small
	\bibliography{egbib}
}
\end{document}